\documentclass[letterpaper, 10 pt, journal, twoside]{IEEEtran}

\IEEEoverridecommandlockouts                              

\usepackage[table]{xcolor}
\usepackage[normalem]{ulem}
\usepackage{colortbl}
\usepackage{changes}
\usepackage{pifont}
\usepackage{wrapfig}
\usepackage{etoolbox}
\usepackage{multirow}
\usepackage{mathtools}
\usepackage{algpseudocode}
\usepackage{xspace}
\usepackage{etoolbox}
\usepackage{pifont}  
\usepackage{graphicx}
\usepackage{epsfig} 
\usepackage{amsmath} 
\usepackage{amssymb}  
\DeclareMathAlphabet{\altmathcal}{OMS}{cmsy}{m}{n}
\usepackage{mathrsfs}
\usepackage{bm}
\usepackage{float}
\usepackage{bbm}
\usepackage{color}
\usepackage{lipsum}
\usepackage{placeins}
\usepackage{afterpage}
\usepackage[export]{adjustbox} 

\definecolor{green1}{rgb}{0.0,0.8,0.4}
\newif\ifshowedits
\showeditsfalse 

\newcommand{\redout}[1]{%
  \ifshowedits
    \textcolor{red}{\sout{#1}}%
  \fi
}

\newcommand{\blu}[1]{%
  \ifshowedits
    \textcolor{blue}{#1}%
  \else
    #1%
  \fi
}

\newif\ifshowmarginnotes
\showmarginnotesfalse 

\newcommand{\marginnote}[1]{%
  \ifshowmarginnotes
    \marginpar{#1}%
  \fi
}

\definecolor{cellred}{RGB}{255, 199, 206}
\definecolor{cellorange}{RGB}{255, 225, 140} 
\definecolor{cellyellow}{RGB}{255, 255, 153}
\usepackage[pagebackref,breaklinks,colorlinks]{hyperref}

\usepackage[capitalize]{cleveref}
    \crefname{section}{Sec.}{Secs.}
    \Crefname{section}{Section}{Sections}
    \Crefname{table}{Table}{Tables}
    \crefname{table}{Tab.}{Tabs.}

\title{SonarSplat: Novel View Synthesis of Imaging Sonar via Gaussian Splatting}

\author{
\begin{tabular}{ccc}
 Advaith V. Sethuraman$^{1}$ & Max Rucker$^{1}$ & Onur Bagoren$^{1}$ \\
 Pou-Chun Kung$^{1}$ & Nibarkavi N.B. Amutha$^{1}$ & Katherine A. Skinner$^{1}$ \\
 \end{tabular} \\
 \vspace{-6mm}
\thanks{Manuscript received: May 4, 2025; Revised August 10, 2025; Accepted October 11, 2025.}
\thanks{This paper was recommended for publication by Editor Sven Behnke upon evaluation of the Associate Editor and Reviewers' comments.}
\thanks{This work was supported by the National Science Foundation under Award No. 2337774.} 
\thanks{$^{1}$Authors are with the Department of Robotics, University of Michigan, Ann Arbor, MI, USA.
        {\tt\footnotesize \{advaiths, mruck, obagoren, pckung, nibah, kskin\}@umich.edu}}%
\thanks{Digital Object Identifier (DOI): see top of this page.}
}

\begin{document}
\maketitle
\begin{abstract}
In this paper, we present SonarSplat, a novel Gaussian splatting framework for imaging sonar that demonstrates realistic novel view synthesis and models acoustic streaking phenomena. Our method represents the scene as a set of 3D Gaussians with acoustic reflectance and saturation properties. We develop a novel method to efficiently rasterize Gaussians to produce a range/azimuth image that is faithful to the acoustic image formation model of imaging sonar. In particular, we develop a novel approach to model azimuth streaking in a Gaussian splatting framework. We evaluate SonarSplat using real-world datasets of sonar images collected from an underwater robotic platform in a controlled test tank and in a real-world river environment. Compared to the state-of-the-art, SonarSplat offers improved image synthesis capabilities (+3.2 dB PSNR) and more accurate 3D reconstruction (\redout{52}\blu{77}\% lower Chamfer Distance). We also demonstrate that SonarSplat can be leveraged for azimuth streak removal.
\end{abstract}

\section{Introduction}
\label{sec:intro}
\IEEEPARstart{A}{coustic} sensors, such as imaging sonar, are commonly used for infrastructure inspection, large-area mapping, and target detection in underwater environments \cite{johannsson2010imaging, ai4shipwrecks, ortho_sonar_englot}. \blu{While optical sensors} \redout{Unlike optical sensors, which} are severely range-limited due to water column effects on light propagation, acoustic sensors can capture data at long ranges to provide critical information about subsea environments. 
\redout{Although sonar exhibits many desirable qualities, such as a longer range, invariance to lighting conditions, and the ability to discern material properties}\blu{However, acoustic phenomena like}\redout{, there exist acoustic phenomena, including} elevation ambiguity, azimuth streaking, and multi-path reflections\redout{, which} make sonar interpretation difficult for both operators and computer vision algorithms.  

\begin{figure}[h!]
\centering
\includegraphics[width=0.9\linewidth]{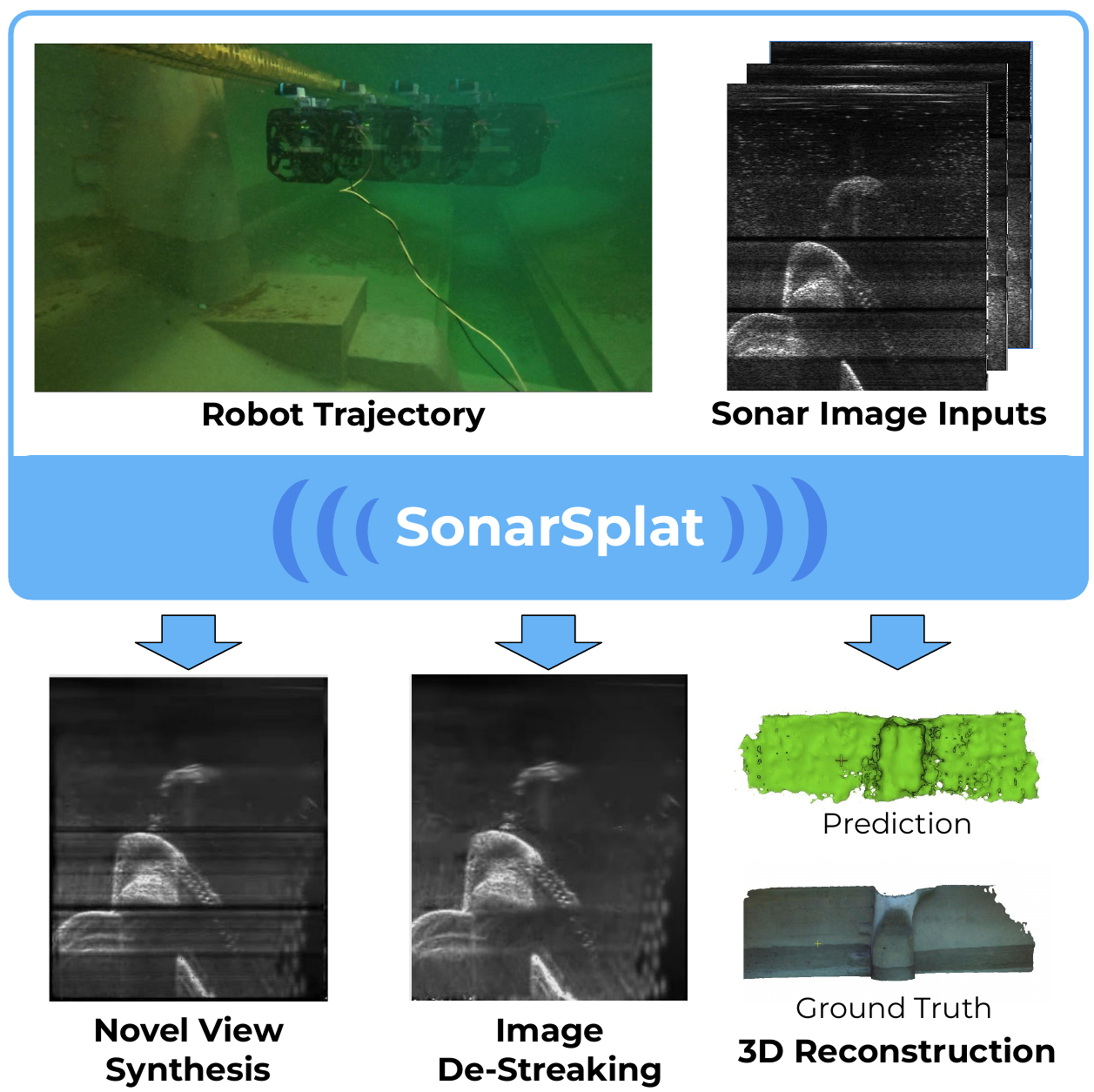}
\caption{We present SonarSplat, a novel 3D Gaussian splatting framework that enables novel view synthesis of realistic sonar images, image de-streaking, and 3D reconstruction. We model the acoustic reflectivity and azimuth streak probabilities for all Gaussians, yielding a 3D scene representation that can be queried for a variety of downstream tasks. 
}
\label{fig:teaser}
\vspace{-7mm}
\end{figure}
Radiance field methods offer a promising approach to addressing these challenges. Neural radiance fields (NeRFs) have demonstrated the potential for high-fidelity data synthesis, denoising, and 3D reconstruction using optical imagery \cite{mildenhall_NeRFRepresentingScenes_2020, NEURIPS2021_e41e164f, pearl2022noiseaware, Chen2024dehazenerf}. The benefits of neural rendering have been recognized by the underwater perception community, with prior work exploring neural rendering for 3D object reconstruction using underwater cameras \cite{levy_SeaThruNeRFNeuralRadiance_2023, sethuraman_WaterNeRFNeuralRadiance_2023} and sonar data \cite{qadri2022neural, qadri2024aoneus, neusis-ngp, dsc, yiping_neural1}. While these innovations are promising, training and processing NeRFs is costly and time-consuming, making deploying them in real-time or on resource-constrained devices difficult. More recently, 3D Gaussian splatting (3DGS) \cite{kerbl3Dgaussians} was developed as a faster alternative to NeRF. 
Gaussian splatting has been leveraged for underwater applications as well \cite{yang2024seasplat, li2024watersplatting, mualem2024gaussiansplashing}. Most relevant to our proposed work, ZSplat proposes a Gaussian splatting framework for RGB-sonar fusion \cite{qu2024zsplatzaxisgaussiansplatting}. However, ZSplat leverages the fusion of sonar data to improve the rendering of RGB images and does not focus on enabling high-fidelity data synthesis for sonar imagery or providing evaluation for the quality of rendered sonar images or 3D reconstruction. Thus, there is a clear gap in the literature for a framework capable of efficient and effective sonar image synthesis and 3D reconstruction using Gaussian splatting. 

In this paper, we propose SonarSplat (\cref{fig:teaser}), a novel Gaussian splatting framework for imaging sonar that enables efficient and high-quality novel view synthesis, sonar image de-streaking, and 3D reconstruction for underwater robotic applications. To the best of our knowledge, SonarSplat is the first sonar-only method that leverages Gaussian splatting to perform these tasks. 
Our main contributions are \redout{summarized} as follows:

\begin{itemize}
    \item We develop a novel 3D Gaussian splatting framework for rasterization of range/azimuth sonar images from known and arbitrary viewpoints.
    \item We develop a novel method for learning the probability of azimuth streaking in a differentiable manner, which allows for removal of unwanted azimuth streaking artifacts from rendered images. 
    \item We introduce a novel densification strategy called the Elevation Sampling Densification Strategy (ESDS) that places Gaussians on elevation arcs and yields better novel view synthesis and 3D reconstruction. 
    \item We perform quantitative and qualitative evaluation of our method on real imaging sonar data collected from a robotic platform with a variety of sonar ranges and environments.
\end{itemize}

\noindent For further information, please visit our project website at \url{https://umfieldrobotics.github.io/sonarsplat3D/}\redout{here}.

\section{Related Work}
\vspace{1mm}
\label{sec: prior_work}



\begin{figure*}[h!]
\vspace{2mm}
\centering
\includegraphics[width=0.85\linewidth]{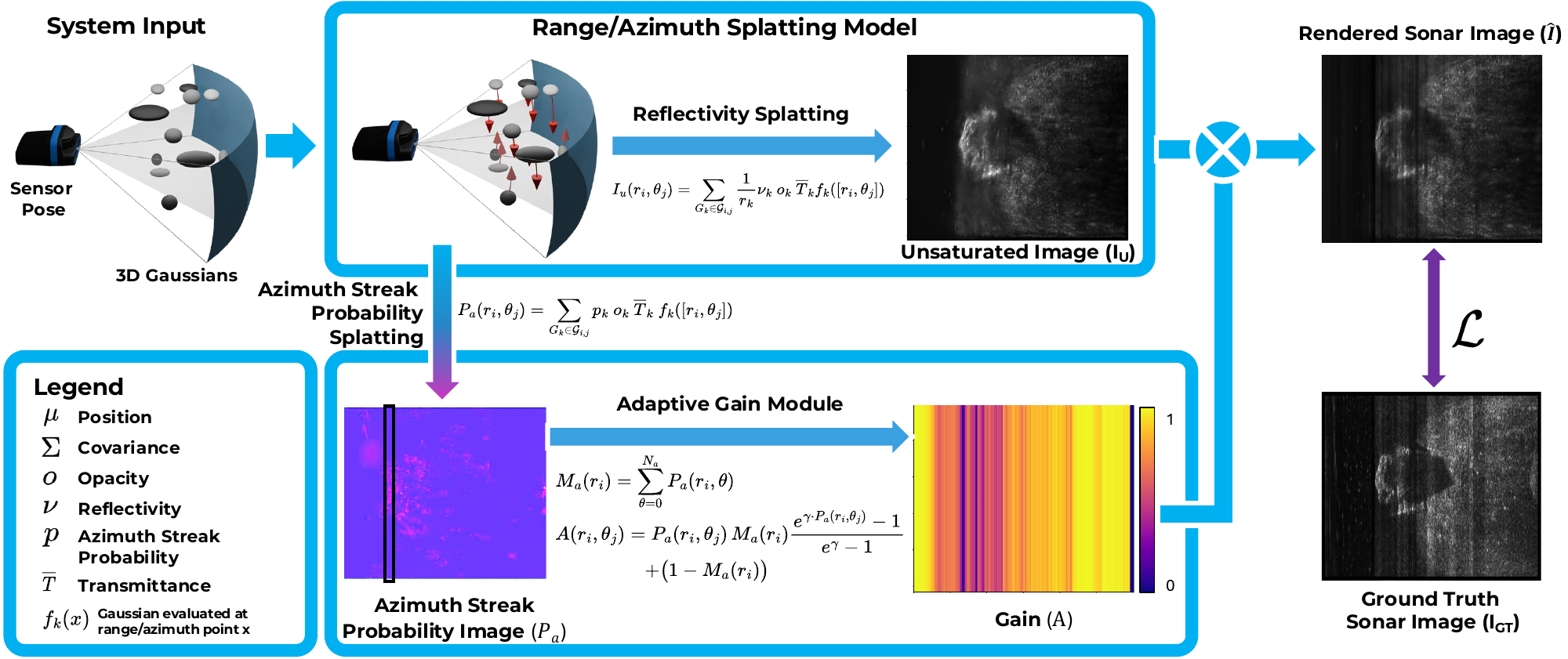}
\caption{Overview of SonarSplat. Our method takes as input a sensor pose and an initial set of 3D Gaussians representing the scene. Then, we transform the 3D Gaussians into the sensor's image space. We splat the reflectivity parameter $\nu_k$ to get the unsaturated image $I_u$.
Additionally, we optimize and splat per-Gaussian azimuth streaking probabilities $p_k$. All of the probabilities in a range bin $r_i$ are considered in our novel Adaptive Gain Module, which adjusts the \redout{receiver} gain \blu{$A$} applied to $I_u$. Finally, we produce $\hat{I}$ by multiplying the gain $A$ by $I_u$. All parameters are optimized using gradient descent by taking losses with respect to the sonar image pixel values. All sonar images shown are polar (range/azimuth) coordinates.
\label{fig:tech_approach}}
\vspace{-5mm}
\end{figure*}

\subsection{Radiance Field Methods}
NeRFs represent scenes implicitly using neural networks and have been explored for a variety of robotic applications, including novel view synthesis \blu{\cite{mildenhall_NeRFRepresentingScenes_2020}}, data generation \blu{\cite{ge2022neural}}, Simultaneous Localization and Mapping (SLAM) \blu{\cite{loner, rosinol_NeRFSLAMRealTimeDense_2022}}, manipulation \blu{\cite{ndf}},\blu{ and} image dehazing \blu{\cite{Chen2024dehazenerf}}\redout{, and trajectory optimization \blu{\cite{9712211}}}\redout{\cite{ge2022neural,loner, rosinol_NeRFSLAMRealTimeDense_2022, ndf, Chen2024dehazenerf}}.
Despite their success as a scene representation, a key challenge for the practical application of NeRFs is the long training time and the slow rendering speeds, primarily due to the computationally expensive volumetric rendering process \cite{mildenhall_NeRFRepresentingScenes_2020}. 
A recent development in radiance field methods is Gaussian splatting, which uses more efficient \textit{rasterization} instead of volumetric rendering to \redout{optimize and }produce a scene representation\blu{. This has led to the use of radiance field methods}\redout{ enabling usage} in a variety of real-time applications \cite{kerbl3Dgaussians, Matsuki:Murai:etal:CVPR2024, luiten2023dynamic}. 

Despite their impressive performance and rapid adoption, radiance field methods were developed with optical sensors in mind, limiting their direct application to non-visual sensors. Recent developments have integrated\redout{ the physics-based image formation models of various} non-visual sensors, including radar, LIDAR, sonar, thermal cameras, and event cameras into NeRF and Gaussian splatting frameworks for various vision tasks \cite{radarfields, cloner, loner, huang2024dartimplicitdopplertomography, qadri2022neural, lin2024thermalnerf}. We are inspired by the faster rendering speeds and high fidelity of Gaussian splatting methods and propose a novel Gaussian splatting framework for imaging sonar.

\subsection{Perception Challenges for Imaging Sonar}
Imaging sonar is a popular underwater sensor for perception tasks like object detection, segmentation, reconstruction, and inspection \cite{qadri2022neural, wide-apeture-reconstruction, johannsson2010imaging, valdenegro2016submerged, stereo_FS}. However, the sensor is susceptible to multi-path effects, nonlinear noise patterns, and phenomena like azimuth streaking \redout{\cite{holoocean_icra}}\blu{\cite{Potokar22iros}}. Azimuth streaking, in particular, can make sonar image understanding difficult by\redout{ suppressing returns from certain azimuth angles}\blu{ suppressing returns from specific ranges\marginnote{\#1.1}}, causing black bars in the sonar image \redout{\cite{holoocean_icra}}\blu{\cite{Potokar22iros}}.
In addition, the sensor inherently loses information from 3D into 2D when capturing the acoustic returns, causing an effect known as \textit{elevation angle ambiguity} \cite{wang20242dforwardlookingsonar}. This ambiguity has made the successful application of imaging sonar to 3D vision tasks challenging \cite{stereo_FS, afm_kaess, albedo_kaess}.

\subsection{Novel View Synthesis for Imaging Sonar}
Prior work has incorporated the sonar image formation model into signed distance field (SDF) and NeRF formulations \cite{qadri2022neural, dsc, neusis-ngp}. Neusis presents a volumetric rendering framework for dense 3D reconstruction of objects using an imaging sonar \cite{qadri2022neural}. 
Neusis-NGP  is an extension of Neusis that leverages multi-resolution hash encodings to produce a bathymetric heightmap of the environment \cite{neusis-ngp}. Differentiable Space Carving (DSC) proposes a faster method that models echo probabilities to carve out a mesh of the scene \cite{dsc}. \redout{Through our experiments, we show that none of these
method are able to perform realistic looking novel view
synthesis for imaging sonar, limiting their application to data generation and augmentation. Furthermore, querying these methods is time consuming and limits their application to real-time methods.}\blu{Through our experiments, we demonstrate that these methods render less realistic sonar imagery compared to SonarSplat according to quantitative metrics including PSNR, SSIM, and LPIPS. This limits their use in data generation and augmentation. Additionally, our experiments indicate that querying these methods is time consuming, which limits their use in real-time applications. \marginnote{\#1.2, \#1.3}}

To the best of our knowledge, ZSplat is the only Gaussian splatting paper that incorporates imaging sonar data, with demonstrations on underwater data. However, \blu{ZSplat operates with a simplified sonar model that does not model several important sonar-specific phenomena.}\redout{ ZSplat oversimplifies the splatting process by not accounting for any sonar-specific phenomena.} More importantly, ZSplat only splats the \textit{opacities} of Gaussians and does not consider material properties like acoustic reflectance. This leads to poor sonar renderings, which limits its application \redout{to}\blu{in} sonar novel view synthesis. We demonstrate that SonarSplat, which more accurately models sonar image formation, produces more realistic novel view synthesis, restores sonar images to remove azimuth streaking artifacts, and produces 3D reconstructions under a unified framework.

\section{Technical Approach}
Figure \ref{fig:tech_approach} shows an overview of SonarSplat, which uses sonar images and their corresponding sensor poses to optimize a scene representation. 
The scene is represented as a set of 3D Gaussian primitives, each parametrized with a mean, covariance, opacity, acoustic reflectivity, and azimuth streaking probability. First, the Gaussians that fall within the view frustum of the sonar are rasterized into a range/azimuth image. 
Concurrently, we render an azimuth streak probability image using the per-Gaussian azimuth streak probabilities. 
Based on the relative probabilities in each range bin, we use a novel adaptive gain module to adjust the acoustic returns and capture azimuth streaking effects. 
Finally, we produce a sonar image with azimuth streaks that can be optimized by taking a loss with respect to the ground truth sonar image. We provide further detail\blu{s} of each of these components in the remainder of this section.

\subsection{Imaging Sonar Formation Model}
To adapt the optical Gaussian splatting framework \redout{to accommodate}\blu{for} imaging sonar, we must incorporate the acoustic image formation model \blu{into the rasterization process}. Imaging sonar is a time-of-flight sensor that\redout{ can} provide\blu{s} the echo intensities at a given range and azimuth bin. \redout{Importantly, imaging sonar\redout{ is not able to} distinguish the elevation angle of a return, leading to an elevation angle ambiguity. This ambiguity makes direct use of imaging sonar difficult for 3D reconstruction and triangulation tasks \cite{Westman19iros, afm_kaess}.}

A sonar image is defined as $I \in \mathbb{R}^{N_a \times N_r}$ where $N_a$ denotes the number of azimuth bins and $N_r$ denotes the number of range bins. 
In spherical coordinates, the sonar has a vertical field of view ($\phi_{min}, \phi_{max}$), a horizontal field of view ($\theta_{min}, \theta_{max}$), and a maximum range, $R_{max}$. Finally,\redout{ taking} $\frac{R_{max}}{N_r}$ \redout{gives us}\blu{is} the range resolution, $\epsilon_r$, and\redout{ taking} $\frac{\theta_{max} - \theta_{min}}{N_a}$ \redout{gives us}\blu{is} the azimuthal resolution, $\epsilon_a$. 

We leverage the sonar rendering equation introduced in \cite{qadri2022neural}, which calculates the intensity of return at bin $(r_i, \theta_j)$ in the sonar image $I$ by: 
\begin{equation}
\resizebox{.95\hsize}{!}{$I(r_i, \theta_j) = \int_{\phi_{min}}^{\phi_{max}} \int_{\theta_j - \epsilon_a}^{\theta_j + \epsilon_a}\int_{r_i - \epsilon_r}^{r_i + \epsilon_r}\frac{E_e}{r}T(r, \theta, \phi)\sigma(r, \theta, \phi)dr d\theta d\phi$} \label{eq:sonar_integral}
\end{equation}
where $E_e$ is the intensity of emitted sound, $r$ is the range of the point, $T(r, \theta, \phi)$ is the transmittance term, and $\sigma(r, \theta, \phi)$ is the density of the point. \redout{Note that this transmittance term}\blu{$T(r, \theta, \phi)$} is\redout{equivalent to the term used in volumetric rendering, serving as} the probability that a ray travels to point $(r, \theta, \phi)$ without hitting another particle. 

\subsection{Range/Azimuth Splatting Model}
\label{sec:ra_splatting}
To evaluate this integral, prior methods use sampling and quadrature \cite{qadri2022neural, qadri2024aoneus, dsc}. 
However,  motivated by the increased efficiency of splatting techniques, we explore a point-based approach adapted to the sonar image formation model.
To this end, we represent the scene with a set of 3D Gaussian primitives $\mathcal{G} = \{G_k\}_{k=1}^N$. \redout{We augment this parametrization with acoustic-specific variables: acoustic reflectivity $\nu_k$ and azimuth streak probability $p_k$.}Formally, these Gaussians are parametrized with a mean $\mu_k \in \mathbb{R}^3$, scale $s_k \in \mathbb{R}^3$, orientation $q_k \in \mathbb{H}$, opacity $o_k \in \mathbb{R}^+$, and \blu{azimuth streak probability $p_k \in \mathbb{R}^+$.} \redout{acoustic reflectivity} \blu{We take inspiration from other Time-of-Flight novel view synthesis work and use spherical harmonics coefficients $sh_k$ to model view-dependent acoustic reflectivity $\nu_k$ \cite{huang2024dartimplicitdopplertomography}. Specifically, $\nu_k = SH(sh_k, \vec{v})$, where $\vec{v}$ is the viewing direction\marginnote{\#1.5, \#2.6}}\redout{$\nu_k \in \mathbb{R}^+$}.
\begin{equation}
    G_k = \redout{\{}\blu{(}\mu_k, s_k, q_k, o_k, \redout{\nu_k,} \blu{sh_k}, p_k\blu{)}\redout{\}}
\end{equation}
where the covariance $\Sigma_k$ \blu{$= R_k S_k S_k^T R_k^T$ for rotation matrix $R_k$ given by quaternion $q_k$ and scaling matrix $S_k$ given by $s_k$.}\redout{ is computed as in \cite{kerbl3Dgaussians}.}

To render sonar images, we first transform the means and covariances of the Gaussians into spherical coordinates. 
Let $\mu_k = \redout{\langle}\blu{[} x_k, y_k, z_k \blu{]^T}\redout{\rangle} ~\text{and}~\Sigma_k$ be the mean and covariance in the sensor's frame, respectively. The transformation to spherical coordinates is as follows \cite{lihigs}:
\begin{equation} 
\mu_s = 
\begin{bmatrix}
    r_k \\
    \theta_k \\ 
    \phi_k 
\end{bmatrix} = 
\begin{bmatrix}
||\mu_k|| \\ 
\arctan(y_k, x_k) \\ 
\arctan(z_k, \sqrt{x_k^2 + y_k^2})\\
\end{bmatrix}
\end{equation}
The covariance matrix $\Sigma_k$ is transformed by using a first-order linearization of the coordinate transformation, which the Jacobian describes:
\begin{equation}
J_s = \begin{bmatrix}
\frac{x_k}{\lVert \mu_k \rVert} & \frac{y_k}{\lVert \mu_k \rVert} & \frac{z_k}{\lVert \mu_k \rVert} \\[1ex]
\frac{-y_k}{x_k^2 + y_k^2} & \frac{x_k}{x_k^2 + y_k^2} & 0 \\[1ex]
\frac{-x_k \cdot z_k}{||\mu_k||^2\sqrt{x_k^2 + y_k^2}} & \frac{-y_k \cdot z_k}{||\mu_k||^2\sqrt{x_k^2 + y_k^2}} & \frac{\sqrt{x_k^2 + y_k^2}}{\lVert \mu_k \rVert^2}
\end{bmatrix}
\end{equation}
and the covariance matrix becomes 
\begin{equation}
\Sigma_s = J_s \Sigma_k J_s^T
\end{equation}

To convert the spherical coordinate mean and covariance into range/azimuth image space, we first specify the intrinsic matrix $K$ of the sonar, 
\begin{equation}
K = 
\begin{bmatrix}
    \frac{1}{\epsilon_r} & 0 & 0 \\
 0 & \frac{1}{\epsilon_a} & \frac{N_a}{2} \\ 
 0 & 0  & 1 \\ 
\end{bmatrix}
\end{equation}
and then we perform the transformation:
\begin{gather}
\mu_k' = K\mu_s,\ \ \Sigma_k' = J_KW \Sigma_s W J_K^T
\end{gather}
where $\mu_k'$ and $\Sigma_k'$ are the mean and covariance transformed to range/azimuth space, $J_K$ is the Jacobian of $K$, and $W$ is the sensor's view matrix, similar to the camera splatting formulation \cite{kerbl3Dgaussians}. 

To find the intensity of acoustic returns at a given range/azimuth bin $[r_i, \theta_j]$, we consider $\mathcal{G}_{i, j} \subset \mathcal{G}$, the set of Gaussians transformed to range/azimuth image space that overlap with the range/azimuth bin $[r_i, \theta_j]$. 
The rasterization equation to obtain the sonar image for SonarSplat is similar to that of \cite{kerbl3Dgaussians}, with a few key differences. \blu{Specifically, we use single-channel acoustic reflectance $\nu$ instead of per-channel color $c$ and we form the image by splatting in the range/azimuth plane using $[r, \theta]$ coordinates instead of traditional camera coordinates. \marginnote{\#2.2}} SonarSplat's rasterization equation is given by: 

\begin{equation}
I_u(r_i, \theta_j) = \redout{\frac{1}{R_k}}\sum_{\redout{G'_k} \blu{G_k} \in \mathcal{G}_{i, j}} \blu{\frac{1}{r_k}}\nu_k~o_k~\blu{\overline{T}}\redout{T}_k~\redout{G'_k(r_i, \theta_j)} \blu{f_k([r_i, \theta_j]) }
\label{eq:raster}
\end{equation}
\blu{
\begin{equation}
f_k(x) = \text{exp}(-\frac{1}{2}\lVert x - \mu'_k \rVert^2_{\Sigma'_k})
\end{equation}
}

\blu{
\begin{equation}
\overline{T}_k = \frac{1}{|\mathcal{P}_k|}\sum_{p \in \mathcal{P}_k}T^p_k
\end{equation}
}

\noindent where \redout{$R_k$}\blu{$r_k$} is the range \blu{of $G_k$} from the sensor,\marginnote{\#2.3} \blu{$f_k(x)$ is the evaluation of Gaussian $G_k$ at range/azimuth point $[r_i, \theta_j]$ with the norm as the Mahalanobis distance,} \blu{$\overline{T}_k$ is the average transmittance of the Gaussian from the sensor origin to the Gaussian along the range.} \redout{and $G'_k$ is the Gaussian evaluated in range/azimuth space.}  \blu{To compute $\overline{T}_k$, 3D Gaussians are splatted onto the azimuth/elevation plane to model occlusions in the range dimension. We define $\mathcal{P}_k$ as the set of azimuth/elevation pixels that the Gaussian $G_k$ overlaps and we calculate $T^p_k$ as the transmittance of $G_k$ at pixel $p$ according to \cite{kerbl3Dgaussians}.\marginnote{\#1.7}}
The subscript indicates that $I_u$ is the \textit{unsaturated} image, meaning that it does not account for the azimuth streaking that causes saturation across the image.
\redout{We observe that the acoustic reflectance of a set of points depends on material properties and viewing angle, $\nu_k$ uses spherical harmonics to encode view-dependent effects.} \redout{In practice, o}\blu{O}ur sonar images are gain adjusted, so \redout{$\frac{1}{R_k}$}\blu{$\frac{1}{r_k}$} is dropped. 


\subsection{Azimuth Streak Modeling (ASM)}
\begin{figure*}[t]
\vspace{3mm}
\centering
\includegraphics[width=0.9\linewidth]{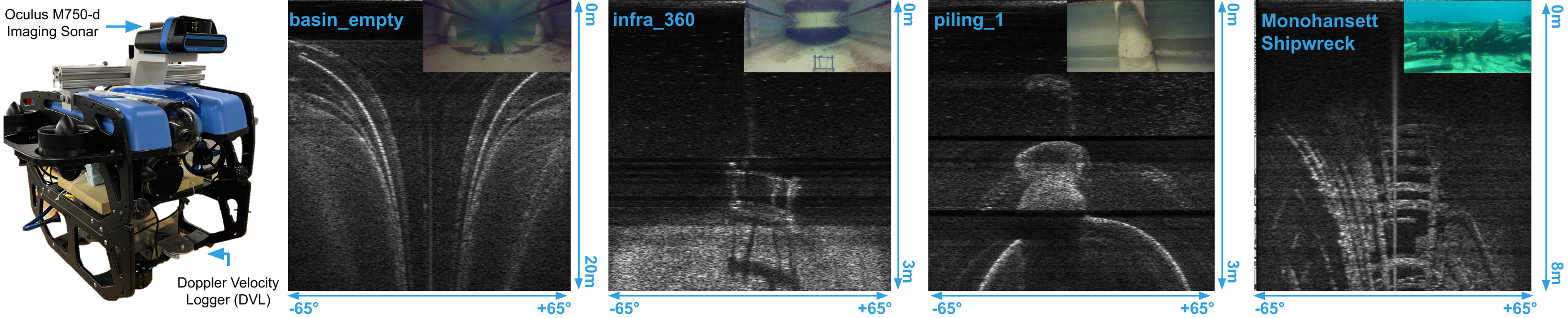}
\caption{We present our robotic platform and selected images from our diverse set of \redout{8}\blu{9} sequences used in our experiments. We show RGB images in top right corners for visualization. Our evaluation datasets focus on smaller objects and larger-scale structures such as the test-tank basin. Azimuth streaks clearly occur in the \texttt{infra\_360} and \texttt{piling\_1} sequences. \redout{Images are range/azimuth.}\label{fig:dset_info}} 
\vspace{-4mm}
\end{figure*}

Azimuth streaking, as shown in \cref{fig:dset_info}, is a phenomena that is frequently observed in sonar images. \redout{It is a form of saturation that occurs when the sonar receiver receives strong returns from incident angles close to parallel at a specific range bin $r_i$ \cite{Potokar22iros}.} \blu{It is a form of saturation that occurs when the sonar receiver receives strong returns from incident angles close to 0 at a specific range bin $r_i$ \cite{Potokar22iros}. \marginnote{\#1.4}} In our model, we assign each Gaussian a probability that it will cause an azimuth streak. 
The benefit of defining this probability per-Gaussian is that it can be optimized to produce azimuth streaks that are multi-view consistent. Let $p_k$ be the probability that Gaussian $k$ will contribute to an azimuth streaking phenomena. 
We can then splat these per-Gaussian probabilities into a range/azimuth bin to obtain an azimuth streak probability image $P_a$:
\begin{gather}
P_a(r_i, \theta_j) = \sum_{\redout{G'_k} \blu{G_k} \in \mathcal{G}_{i, j}} p_k~o_k~\blu{\overline{T}}\redout{T}_k~\redout{G'_k(r_i, \theta_j)} \blu{f_k([r_i, \theta_j])} \\
M_a(r_i) = \sum_{\theta=0}^{N_a} P_a(r_i, \theta)
\end{gather}
where $M_a(r_i)$ represents the probability of an azimuth streak occurring at range interval $r_i$. 
The azimuth streak probabilities across the range bins are then used to compute the final image using an adaptive gain mechanism.

\subsubsection{Adaptive Gain Module}
We introduce a novel adaptive gain term, $A$, \blu{\marginnote{\#2.4} that is produced by transforming the azimuth streak probability image}\redout{to transform the azimuth streak probability image into receiver gains}. 
We begin designing this gain term with a few observations. 
First, when no Gaussian in a range bin $r_i$ has a high probability of azimuth streaking, the gain should be unity. 
Second, suppose a single Gaussian $G_k$ exhibits a high probability of azimuth streaking across the range bin $r_i$. 
In that case, we wish to adaptively \textit{suppress} the other Gaussians in $r_i$ and assign a high gain to $G_k$. 
Finally, if multiple Gaussians in $r_i$ have high probabilities of contributing to an azimuth streak, higher gain should be assigned to those Gaussians.
Following these observations, the adaptive gain term is generated for each pixel: $(r_i, \theta_j)$:
 \begin{equation}
\resizebox{.95\hsize}{!}{$A(r_i, \theta_j) =  P_a(r_i, \theta_j) M_a(r_i)\frac{e^{\gamma \cdot P_a(r_i, \theta_j)} - 1}{e^{\gamma} - 1} + (1 - M_a(r_i))$}
 \end{equation}
where $\gamma$ is a scaling factor that dictates the steepness of the adaptive gain.
This behavior aligns with the insight presented in \cite{Potokar22iros}: if a certain percentage of returns exceeds a threshold, an operation is applied to all of the values in a range interval. 
However, our proposed model differs because we do not restrict the gain to a fixed function (quadratic) but rather offer a family of curves for our splatting model to explore during optimization. 
The final sonar image $\hat{I}(r_i, \theta_j)$ with azimuth streaks is then computed by:
\begin{equation}
    \hat{I}(r_i, \theta_j) = A(r_i, \theta_j)\cdot I_u(r_i,\theta_j)
\end{equation}
\begin{figure}[t]
\centering
\includegraphics[width=0.75\linewidth]{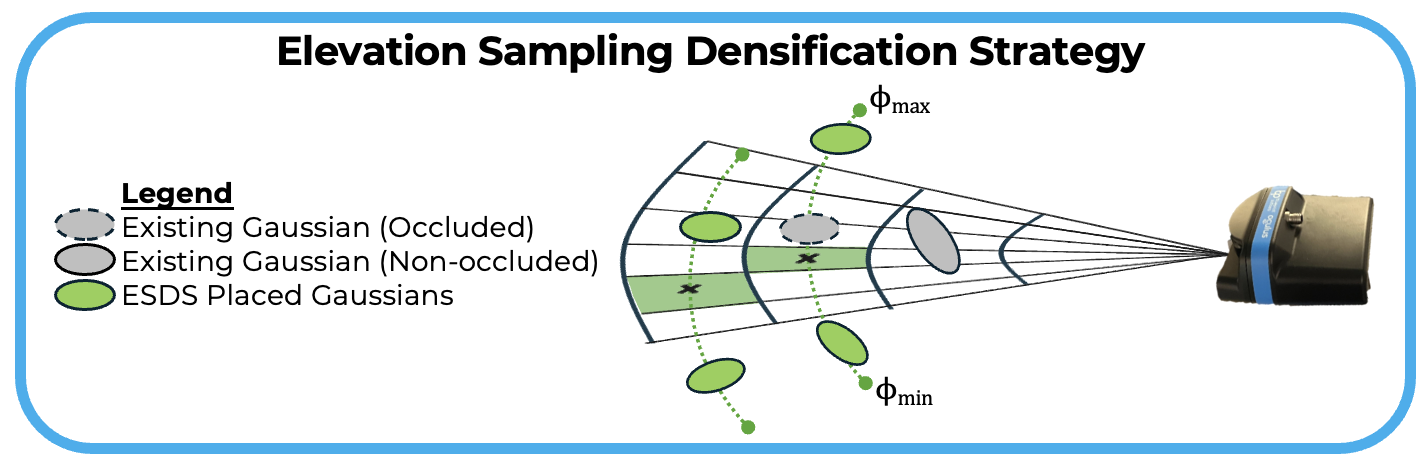}
\caption{Illustration of the proposed Elevation Sampling Densification Strategy (ESDS). We randomly sample $N_p$ pixels (shown as green sectors) based on the magnitude of $\mathcal{L}$ then randomly place $N_g$ Gaussians (shown as green ellipses) on the elevation arc of that pixel. \label{esds_img}}
\vspace{-6mm}
\end{figure}
\vspace{-2mm}
\subsection{Model Optimization}
We optimize \redout{for desired parameters}\blu{SonarSplat} using gradient descent and by taking losses ($\mathcal{L}_{l1}, \mathcal{L}_{ssim}$) between $\hat{I}$ and the ground truth image $I_\text{GT}$ \cite{kerbl3Dgaussians}. Next, we note that large Gaussians cause increased training times and more distortion to the model presented in \cref{sec:ra_splatting}. To prevent Gaussians from becoming too large, we use $\mathcal{L}_{size} = \text{ReLU}(s - s_{init})$, which encourages the scale of the Gaussians $s$ to remain smaller than their initialized scale $s_{init}$ \cite{lee2025microsplattingmaximizingisotropicconstraints}. Additionally, we find it useful for reconstruction tasks to supervise the opacities \redout{in regions with sonar returns}\blu{of Gaussians} via an opacity loss $\mathcal{L}_o$, similar to \cite{qu2024zsplatzaxisgaussiansplatting}. We compute $\mathcal{L}_o$ between the predicted opacity map $\hat{I}_o$ and $I_{\text{GT}}$ thresholded at $\tau_o$ to isolate strong returns. The threshold parameter is found experimentally for each dataset. $\mathcal{L}_o$ then becomes an $l1$ loss between $\hat{I}_o$ and $I^o_{\text{GT}}$. 
The final loss becomes: 
\begin{equation}
\mathcal{L} = \lambda_{l1}\mathcal{L}_{l1} + (1-\lambda_{l1})\mathcal{L}_{ssim} + \lambda_{o}\mathcal{L}_o + \lambda_{size}\mathcal{L}_{size}
\end{equation}
\noindent where $\lambda_{l1}, \lambda_{o},~\text{and}~\lambda_{size}$ are the weights for the $l1$, opacity, and size losses respectively. 

\subsection{Elevation Sampling Densification Strategy (ESDS)}
Splitting and cloning Gaussians in image space is unintuitive for imaging sonar given the elevation angle ambiguity. It is possible that a Gaussian can be split then immediately occluded by high opacity Gaussians in front of it. To address this, we introduce an alternative strategy for densification of Gaussians called the Elevation Sampling Densification Strategy (ESDS) shown in \cref{esds_img}. Given predicted image $\hat{I}$ and ground truth $I_{\text{GT}}$, we sample $N_p$ pixels, where each pixel's probability is proportional to the magnitude of the loss at that pixel. We initialize $N_g$ Gaussians on each elevation arc specified by the sampled pixel locations. This primarily serves to provide SonarSplat enough opportunities to resolve the elevation angle ambiguity and any possible occlusions by placing Gaussians at different elevation angles. \redout{As the optimization continues, l}\blu{L}ow-opacity Gaussians are pruned similar to \cite{kerbl3Dgaussians}. 

\subsection{Implementation Details}
SonarSplat is built on the \texttt{gsplat} repo \cite{ye2024gsplatopensourcelibrarygaussian}. We train our method on an NVIDIA RTX 3090 GPU with 24 GB of vRAM. We calculate the frames per second (FPS) of all methods on the same NVIDIA A6000 GPU with 50 GB of vRAM. \blu{For every pixel in a sonar image with intensity above a threshold $\tau_{init}$, we initialize $N_{init}$ Gaussians on the elevation arc specified by the pixel's range, azimuth, and the sensor's vertical field of view. This process is repeated for all sonar images in a dataset to produce the initial set of 3D Gaussians. \marginnote{\#2.1}}
To properly optimize for the azimuth streaking probabilities $p_k$ of each Gaussian, we first identify where azimuth streaks occur by calculating the average intensity of the range interval $r_i$. 
Then, for the first $N_s$ iterations, we train only on pixels in range bins that exceed a certain average intensity threshold $\tau_s$. This avoids incorrectly supervising the splat in azimuth streaked regions\marginnote{\#2.5}.  
Importantly, $p_k$ is not optimized during this interval. After $N_s$ iterations, we optimize only $p_k$ in order to capture the azimuth streaking effects. We do not optimize $\mu_k, ~\Sigma_k,~o_k$, or $\nu_k$ during this interval. This \redout{way, we }allow\blu{s} the Gaussians to \redout{initialize and }fit the images before optimizing the azimuth streaking parameters.

\section{Results \& Experiments}

\begin{figure*}[t]
\vspace{2mm}
    \centering
    \includegraphics[width=0.85\linewidth]{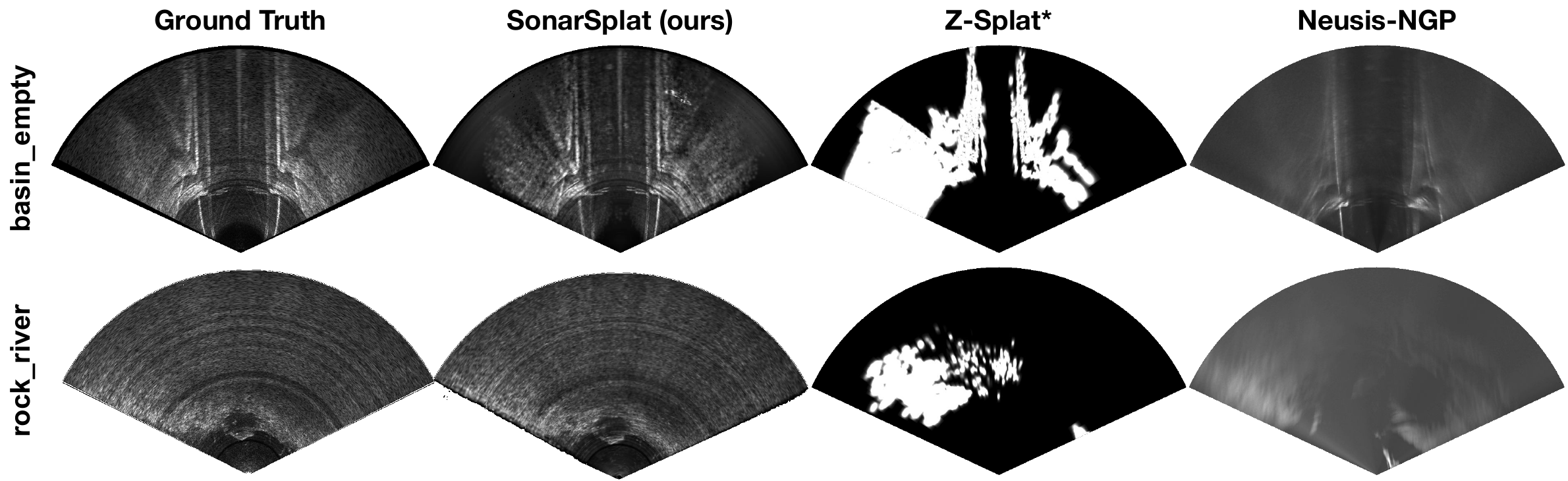}
    \caption{We present qualitative novel view synthesis results from selected datasets and selected baselines. Note that SonarSplat produces more realistic sonar image synthesis compared to baselines and is better able to capture finer details in the environment. Note that ZSplat* indicates ZSplat trained only with the sonar loss. Images are cartesian. \label{fig:qual_results1}}
    \vspace{-5mm}
\end{figure*}

\subsection{Robotic Platform and Datasets}
We note that real, public datasets of posed sonar images suitable for novel view synthesis and reconstruction experiments do not exist. Rather, we collect sequences using a BlueRobotics BlueROV2 platform, shown in \cref{fig:dset_info}. Our robotic platform is equipped with a BluePrint Subsea Oculus M750-d imaging sonar operating at 1.2MHz, capturing sequences between 3-20m in range, with a 130$^\circ$ horizontal field of view and a 20$^\circ$ vertical field of view. \blu{The Oculus M750-d has a range resolution varying from 0.5-4 cm/pixel and angular resolution of 0.6$^\circ$ with a beam separation of 0.25$^\circ$ \cite{subsea_blueprint_nodate}. \marginnote{\#1.8}} We tilt the sonar both at 30$^\circ$ and 0$^\circ$ downwards for various sequences.  We use this platform to collect \redout{8}\blu{9} trajectories inspecting a range of objects and in both a test tank and river environment. We use 7 sequences for evaluating image synthesis, and \redout{1}\blu{2} sequence\blu{s} with higher coverage of the inspected object for evaluating 3D reconstruction. \blu{Our 3D reconstruction datasets are called \texttt{Concrete Piling} and \texttt{\textit{Monohansett} Shipwreck.}} Sensor poses are estimated by filtering a Doppler Velocity Logger and IMU, similar to \cite{noauthor_dvl_nodate, qadri2022neural}. \blu{For \texttt{Concrete Piling},} \redout{W}\blu{w}e construct ground truth for 3D reconstruction using RGB images and MASt3R, scaling the reconstruction to metric scale \cite{leroy2024groundingimagematching3d}.\redout{We find that MASt3R produces less artifacts than traditional photogrammetry pipelines like AgiSoft Metashape \cite{agisoft2025}.} \blu{For \texttt{\textit{Monohansett} Shipwreck}, we use an underwater LiDAR scan of the shipwreck as ground truth.}

\subsection{Baselines}
For baselines, we compare to Neusis~\cite{qadri2022neural}, Neusis-NGP~\cite{neusis-ngp}, DSC~\cite{dsc}, and ZSplat*~\cite{qu2024zsplatzaxisgaussiansplatting}. Since we are focused on sonar-only rendering, we compare to a modified version, ZSplat*, trained with only the sonar loss since there is no other work that has explored Gaussian splatting for imaging sonar \cite{qu2024zsplatzaxisgaussiansplatting}. All baselines were trained using official\redout{released} code and hyperparameters were tuned to the best of our ability.

\begin{table}[t]
\caption{Novel view synthesis performance on real world held-out validation data \blu{(7 datasets)}. Note that ZSplat* is ZSplat with only the sonar loss activated for training. First, second, and third best are shown in red, orange, and yellow\redout{, respectively}. \label{tab:NVS results}}
\centering
\scalebox{0.9}{
\begin{tabular}{lcccc} \hline
\multicolumn{5}{c}{\textbf{Sonar Image Synthesis}} \\ \hline
Method & PSNR $\uparrow$ & SSIM $\uparrow$ & LPIPS $\downarrow$ & FPS $\uparrow$ \\ \hline

Neusis~\cite{qadri2022neural}           
& 14.68 & 0.04 & \cellcolor{cellyellow}0.64 & \cellcolor{cellyellow}0.03 \\

Neusis-NGP~\cite{neusis-ngp}           
& \cellcolor{cellorange}19.56 & \cellcolor{cellorange}0.33 & \cellcolor{cellorange}0.64 & 0.01 \\

DSC~\cite{dsc}                          
& \cellcolor{cellyellow}14.70 & \cellcolor{cellyellow}0.07 & 0.66 & \cellcolor{cellyellow}0.03 \\

ZSplat*~\cite{qu2024zsplatzaxisgaussiansplatting} 
& 10.06 & 0.02 & 0.69 & \cellcolor{cellred}155.32 \\ \hline

SonarSplat (ours)             
& \cellcolor{cellred}22.79 
& \cellcolor{cellred}0.41
& \cellcolor{cellred}0.48 
& \cellcolor{cellorange}13.45 \\ \hline
\end{tabular}
}
\vspace{-4mm}
\end{table}

\subsection{Novel View Synthesis \label{nvs_main}}
 We report validation PSNR, SSIM, and LPIPS following \cite{kerbl3Dgaussians}, using the \texttt{lpipsPyTorch} library. For our validation set, we take every 8th image from the dataset as introduced in~\cite{mildenhall_NeRFRepresentingScenes_2020}. We report SonarSplat's performance compared to baselines in \cref{tab:NVS results}. We find that SonarSplat consistently synthesizes more realistic sonar images, quantified by its higher PSNR and SSIM, and lower LPIPS values. Qualitative results from selected sequences are presented in \cref{fig:qual_results1}. We also report the rendering speed in FPS of each method in \cref{tab:NVS results}. SonarSplat and ZSplat* both leverage Gaussian splatting, which offers significantly faster rendering speeds compared to neural rendering methods. ZSplat* is optimized for \redout{high}\blu{fast} rendering\redout{ speed}, which will be a focus of future work. 

\begin{table}[b]
\vspace{-4mm}
\caption{Ablation study results for the proposed acoustic reflectance parameter (VDR), azimuth streak modeling (ASM), and elevation sampling densification strategy (ESDS). Best in \textbf{bold}. \label{tab:nvs_ablation}}
\centering
\scalebox{0.8}{
\begin{tabular}{cccccc}
\hline
\multicolumn{6}{c}{\textbf{SonarSplat Rendering Ablations}} \\ \hline
\multicolumn{1}{c}{\begin{tabular}[c]{@{}c@{}}VDR\end{tabular}} &
\multicolumn{1}{c}{\begin{tabular}[c]{@{}c@{}}ASM\end{tabular}} &
\multicolumn{1}{c}{\begin{tabular}[c]{@{}c@{}}ESDS\end{tabular}} &
\multicolumn{1}{c}{PSNR $\uparrow$} &
\multicolumn{1}{c}{SSIM $\uparrow$} &
\multicolumn{1}{c}{LPIPS $\downarrow$} \\ \hline
\ding{55} & \ding{55} & \ding{55} & 21.32 & 0.35 & 0.53 \\
\ding{51} & \ding{55} & \ding{55} & 21.16 & 0.34 & 0.55 \\
\ding{51} & \ding{51} & \ding{55} & 22.51 & 0.40 & 0.52 \\ \hline
\ding{51} & \ding{51} & \ding{51} & \textbf{22.79} &\textbf{0.41} & \textbf{0.48} \\ \hline
\end{tabular}
}
\end{table}

\subsection{Ablation Studies}
\begin{figure}[t]
\includegraphics[width=\linewidth]{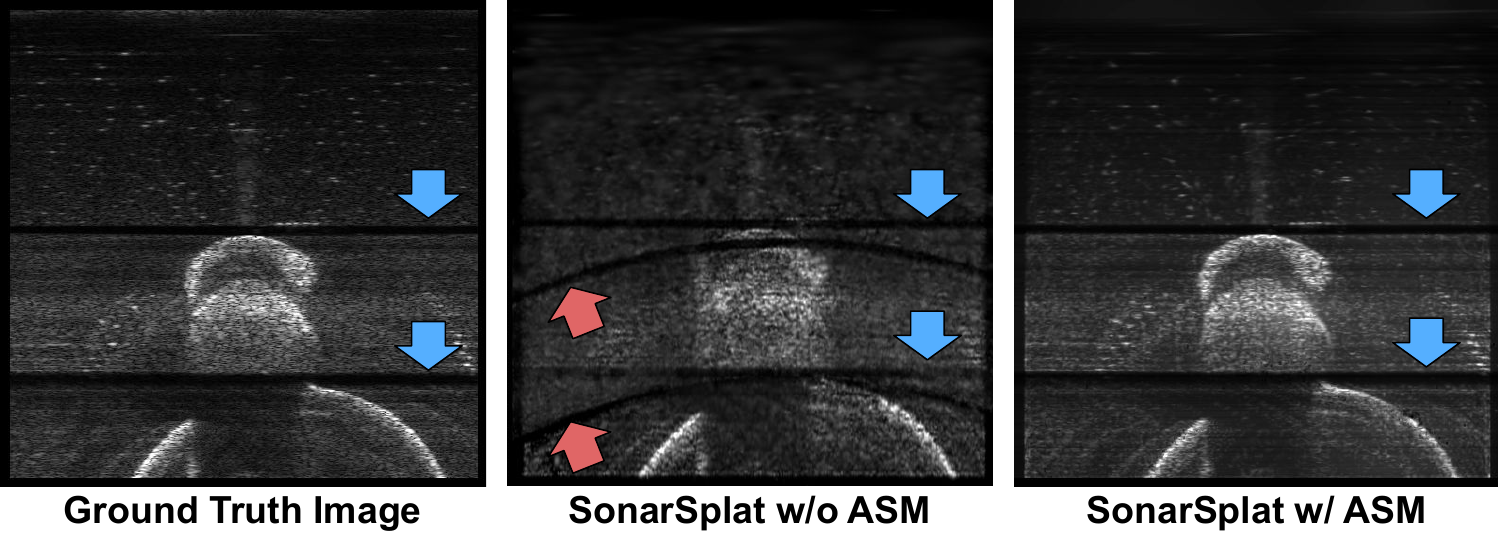}
\caption{Qualitative comparison of SonarSplat without and with azimuth streak modeling (ASM) for \texttt{piling\_1}. Note that if ASM is not modeled, rendered views will produce erroneous streaks (indicated with red arrows) instead of capturing the true streaks (indicated with blue arrows). \label{fig:abl_azimuth}}
\end{figure}

We perform ablation studies on the view-dependent reflectivity parameter $\nu_k$ (VDR), the azimuth streak modeling (ASM), and the Elevation Sampling Densification Strategy (ESDS). Ablation studies are averaged across 7 evaluation datasets and trained with exactly the same hyperparameters as results from \cref{nvs_main}. Methods that don't use ESDS use the \texttt{mcmc} strategy in \texttt{gsplat} \cite{ye2024gsplatopensourcelibrarygaussian}. Results are reported in \cref{tab:nvs_ablation} and shown in \cref{fig:abl_azimuth}. Clearly, the addition of azimuth streak modeling, view-dependent acoustic reflectivity, and the elevation sampling densification strategy improves the image synthesis capabilities of SonarSplat.  

\begin{figure}[t]
\centering
\includegraphics[width=0.8\linewidth]{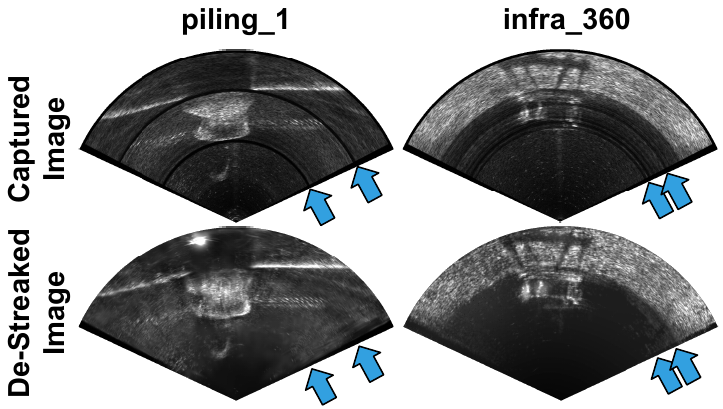}

\caption{Qualitative results for azimuth streak removal capabilities of SonarSplat. Locations of azimuth streaks in the captured images are indicated by blue arrows. \label{fig:streak_removal}
}
\vspace{-3mm}
\end{figure}



\begin{table}[h!]
\caption{\redout{Quantitative results for azimuth streak removal.} We report the inverse coefficient of variation (ICV) \blu{for azimuth streak removal}. \redout{Higher is better ($\uparrow$) and b}\blu{B}est is in \textbf{bold}. \label{tab:streak_removal}}
\centering
\scalebox{0.8}{
\begin{tabular}{lcc} \hline
\multicolumn{3}{c}{\textbf{De-Streaking Performance (ICV) $\uparrow$}} \\ \hline
 & Captured Image & De-Streaked Prediction \\ \hline
Average & 2.91 & \textbf{18.58} \\ \hline
\end{tabular}
}

\vspace{-5mm}
\end{table}

\subsection{Azimuth Streak Removal/De-Streaking}
Using SonarSplat, we can undo the adaptive gain to render images without azimuth streaks present and recover suppressed returns. This vision task is most similar to \textit{image de-striping} from the remote sensing literature. We present quantitative results in \cref{tab:streak_removal} using the inverse coefficient of variation (ICV) metric, which is a common no-reference metric used in de-striping works \cite{shen2008map, rakwatin2007stripe, nichol2004noise}. We present qualitative results of de-streaked images in \cref{fig:streak_removal}. The Azimuth Streak Modeling allows SonarSplat to qualitatively and quantitatively remove streaks effectively and improves the finer features in sonar images.


\begin{figure*}
\vspace{3mm}
\centering
\includegraphics[width=\linewidth]{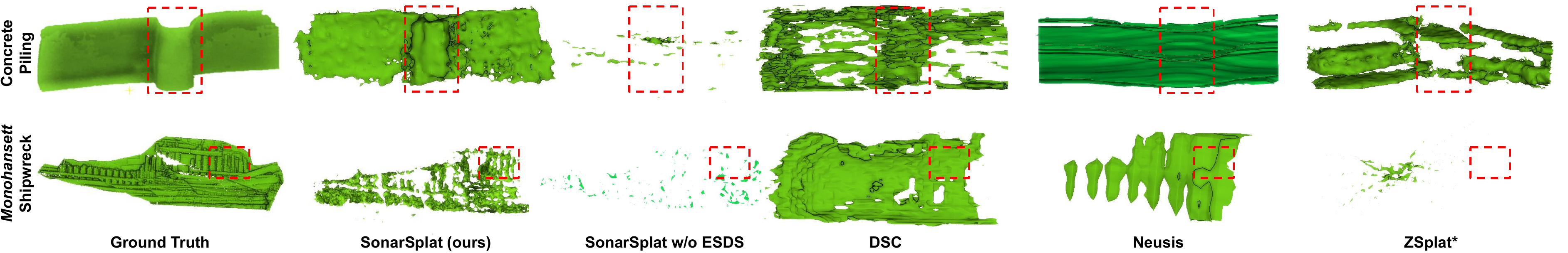}
\caption{Qualitative results from 3D reconstruction of SonarSplat compared to baselines. We inspect \redout{a concrete piling}\blu{\texttt{Concrete Piling} in a test tank. \texttt{\textit{Monohansett} Shipwreck} was surveyed in Lake Huron, Michigan.} Ground truth is created using collected RGB images \blu{and LIDAR respectively. Readers are encouraged to zoom in}. \label{fig:3D_qual}}
\vspace{-6mm}
\end{figure*}
\subsection{Real-World 3D Reconstruction}
3D reconstruction is crucial for inspection and visualization of underwater structures. In a test tank, we demonstrate inspection of \texttt{Concrete Piling}, a structure commonly found in bridges and dams. \blu{We also surveyed \texttt{\textit{Monohansett} Shipwreck}, a wooden shipwreck at a depth of 18 feet in Lake Huron, Michigan, measuring 160 feet in length and 30 feet in width.}

We show that SonarSplat learns the geometric properties of the scene by converting the optimized splat into a mesh for visualization. \blu{We sample a set of points from the probability density functions described by the Gaussians, then query the opacities and use marching cubes to produce a mesh. Specifically, we used between 10-50 samples per Gaussian depending on the size of the scene and the density of the Gaussians.\marginnote{\#2.7, \#2.8, \#2.10}} We compare to state-of-the-art sonar reconstruction methods: Neusis, DSC, and ZSplat* \cite{dsc, qadri2022neural, qu2024zsplatzaxisgaussiansplatting}. \blu{Additionally, we ablate ESDS (denoted as ``SonarSplat w/o ESDS").} Qualitative results are shown in \cref{fig:3D_qual} and quantitative results are shown in \cref{tab:3D_quantitative}. Following \cite{qadri2022neural, dsc}, we report Root Mean Squared (RMS) $l1$ Chamfer Distance (CD-$l1$) and Hausdorff Distance (HD)\blu{, which are both in meters}. We produce meshes from all methods then randomly sample 30,000 points from the surface for 30 iterations to calculate RMS. \blu{We use ICP to align the predictions and ground truth meshes then crop the predictions to the bounding box of the ground truth.} Note that neither DSC, Neusis, nor ZSplat* is able to reconstruct the object whereas SonarSplat is able to produce an accurate reconstruction of the piling. \blu{For \texttt{\textit{Monohansett} Shipwreck}, we see that SonarSplat captures details like the wooden ribs of the shipwreck more accurately than baselines.} 
\begin{table}[h!]
\vspace{-2mm}
\caption{Quantitative evaluation of 3D reconstruction\redout{performance on the piling} using RMS $l1$ Chamfer Distance (CD-$l1$) and Hausdorff Distance (HD). \blu{Units are meters.} Best in \textbf{bold}.}
\centering
\scalebox{0.9}{

\begin{tabular}{lccll|}
\hline
\multicolumn{5}{|c|}{\textbf{3D Reconstruction \redout{Metrics}\blu{Performance}}}                                                                                                                                                                                               \\ \hline
\multicolumn{1}{|l|}{}                                                                   & \multicolumn{2}{c|}{Concrete Piling}                         & \multicolumn{2}{c|}{\blu{\textit{Monohansett} Shipwreck}}                                                 \\ \hline
\multicolumn{1}{|l|}{Method}                                                             & CD-$l_1$ $\downarrow$ & \multicolumn{1}{c|}{HD $\downarrow$} & \multicolumn{1}{c}{\blu{CD-$l_1$ $\downarrow$}} & \multicolumn{1}{c|}{\blu{HD $\downarrow$}} \\ \hline
\multicolumn{1}{|l|}{Neusis \cite{qadri2022neural} }                     & 0.32                  & \multicolumn{1}{c|}{0.78}            &          \multicolumn{1}{c}{\blu{1.75}}                                   &   \multicolumn{1}{c|}{\blu{3.67}}                                    \\
\multicolumn{1}{|l|}{DSC \cite{dsc}}                                    & 0.29                  & \multicolumn{1}{c|}{0.82}            & \multicolumn{1}{c}{\blu{1.60}}                  & \multicolumn{1}{c|}{\blu{6.77}}                  \\
\multicolumn{1}{|l|}{ZSplat* \cite{qu2024zsplatzaxisgaussiansplatting}} & 2.93                  & \multicolumn{1}{c|}{4.88}            &        \multicolumn{1}{c}{\blu{3.06}}                                   &     \multicolumn{1}{c|}{\blu{8.30}}                                   \\ 
\multicolumn{1}{|l|}{\blu{SonarSplat w/o ESDS}}                                                   & \blu{0.92}         & \multicolumn{1}{c|}{\blu{1.77}}   & \multicolumn{1}{c}{\blu{3.62}}                  & \multicolumn{1}{c|}{\blu{12.52}}            \\ \hline

\multicolumn{1}{|l|}{SonarSplat (ours)}                                                   & \textbf{0.14}         & \multicolumn{1}{c|}{\textbf{0.57}}   & \multicolumn{1}{c}{\blu{\textbf{0.29}}}                  & \multicolumn{1}{c|}{\blu{\textbf{1.06}}}            \\ \hline
\end{tabular}

}

\label{tab:3D_quantitative}
\vspace{-3mm}
\end{table}


\section{Conclusion \& Future Work}
We propose SonarSplat, the first sonar-only Gaussian splatting framework for imaging sonar in underwater applications. 
First, we adapt the sonar rendering equation for efficient range/azimuth splatting by evaluating 3D Gaussians. 
We model azimuth streaking within the Gaussian splatting framework, allowing us to learn per-Gaussian azimuth streaking probabilities and produce high-quality de-streaked images. We also introduce a novel densification strategy (ESDS) to enable SonarSplat to resolve elevation angle ambiguities and associated occlusions. From our extensive experiments on a diverse set of real-world data in test tank and lake environments, we find that SonarSplat outperforms state-of-the-art methods in novel view synthesis (+3.2 dB PSNR) and 3D reconstruction (\redout{52}\blu{77}\% lower Chamfer Distance), and additionally allows for de-streaking of sonar images. 

Future work will focus on leveraging SonarSplat for sonar data synthesis by randomizing scene parameters including acoustic reflectance and azimuth streaking probability. This can yield more diverse datasets for training sonar-based scene understanding algorithms, including multi-view target recognition algorithms \cite{Sethuraman_2023_BMVC, gmvatr}. We will also work to improve SonarSplat's ability to perform dense 3D reconstruction by better modeling additional acoustic phenomena like speckle noise\redout{, Lambertian scattering,} and multi-path reflections. Our choice of Gaussian splatting as a 3D scene representation also opens possibilities for real-time sensor fusion and SLAM \cite{Matsuki:Murai:etal:CVPR2024}.

\bibliographystyle{IEEETran}
\bibliography{main}

\begin{thebibliography}{10}
\providecommand{\url}[1]{#1}
\csname url@rmstyle\endcsname
\providecommand{\newblock}{\relax}
\providecommand{\bibinfo}[2]{#2}
\providecommand\BIBentrySTDinterwordspacing{\spaceskip=0pt\relax}
\providecommand\BIBentryALTinterwordstretchfactor{4}
\providecommand\BIBentryALTinterwordspacing{\spaceskip=\fontdimen2\font plus
\BIBentryALTinterwordstretchfactor\fontdimen3\font minus \fontdimen4\font\relax}
\providecommand\BIBforeignlanguage[2]{{%
\expandafter\ifx\csname l@#1\endcsname\relax
\typeout{** WARNING: IEEEtran.bst: No hyphenation pattern has been}%
\typeout{** loaded for the language `#1'. Using the pattern for}%
\typeout{** the default language instead.}%
\else
\language=\csname l@#1\endcsname
\fi
#2}}

\bibitem{johannsson2010imaging}
H.~Johannsson, M.~Kaess, B.~Englot, F.~Hover, and J.~Leonard, ``Imaging sonar-aided navigation for autonomous underwater harbor surveillance,'' in \emph{2010 IEEE/RSJ International Conference on Intelligent Robots and Systems}.\hskip 1em plus 0.5em minus 0.4em\relax IEEE, 2010, pp. 4396--4403.

\bibitem{ai4shipwrecks}
\BIBentryALTinterwordspacing
A.~V. Sethuraman, A.~Sheppard, O.~Bagoren, C.~Pinnow, J.~Anderson, T.~C. Havens, and K.~A. Skinner, ``Machine learning for shipwreck segmentation from side scan sonar imagery: Dataset and benchmark,'' \emph{The International Journal of Robotics Research}, vol.~44, no.~3, pp. 341--354, 2024. [Online]. Available: \url{https://doi.org/10.1177/02783649241266853}
\BIBentrySTDinterwordspacing

\bibitem{ortho_sonar_englot}
J.~McConnell, J.~D. Martin, and B.~Englot, ``Fusing concurrent orthogonal wide-aperture sonar images for dense underwater 3{D} reconstruction,'' in \emph{2020 IEEE/RSJ International Conference on Intelligent Robots and Systems (IROS)}, 2020, pp. 1653--1660.

\bibitem{mildenhall_NeRFRepresentingScenes_2020}
B.~Mildenhall, P.~P. Srinivasan, M.~Tancik, J.~T. Barron, R.~Ramamoorthi, and R.~Ng, ``{{NeRF}}: {{Representing Scenes}} as {{Neural Radiance Fields}} for {{View Synthesis}},'' in \emph{2020 European Conference on Computer Vision (ECCV)}, ser. Lecture {{Notes}} in {{Computer Science}}, A.~Vedaldi, H.~Bischof, T.~Brox, and J.-M. Frahm, Eds.\hskip 1em plus 0.5em minus 0.4em\relax {Cham}: {Springer International Publishing}, 2020, pp. 405--421.

\bibitem{NEURIPS2021_e41e164f}
\BIBentryALTinterwordspacing
P.~Wang, L.~Liu, Y.~Liu, C.~Theobalt, T.~Komura, and W.~Wang, ``Neus: Learning neural implicit surfaces by volume rendering for multi-view reconstruction,'' in \emph{Advances in Neural Information Processing Systems}, M.~Ranzato, A.~Beygelzimer, Y.~Dauphin, P.~Liang, and J.~W. Vaughan, Eds., vol.~34.\hskip 1em plus 0.5em minus 0.4em\relax Curran Associates, Inc., 2021, pp. 27\,171--27\,183. [Online]. Available: \url{https://proceedings.neurips.cc/paper_files/paper/2021/file/e41e164f7485ec4a28741a2d0ea41c74-Paper.pdf}
\BIBentrySTDinterwordspacing

\bibitem{pearl2022noiseaware}
\BIBentryALTinterwordspacing
N.~Pearl, T.~Treibitz, and S.~Korman, ``{ NAN: Noise-Aware NeRFs for Burst-Denoising },'' in \emph{2022 IEEE/CVF Conference on Computer Vision and Pattern Recognition (CVPR)}.\hskip 1em plus 0.5em minus 0.4em\relax Los Alamitos, CA, USA: IEEE Computer Society, June 2022, pp. 12\,662--12\,671. [Online]. Available: \url{https://doi.ieeecomputersociety.org/10.1109/CVPR52688.2022.01234}
\BIBentrySTDinterwordspacing

\bibitem{Chen2024dehazenerf}
W.-T. Chen, W.~Yifan, S.-Y. Kuo, and G.~Wetzstein, ``Dehazenerf: Multi-image haze removal and 3{D} shape reconstruction using neural radiance fields,'' in \emph{2024 International Conference on 3D Vision (3DV)}, 2024, pp. 247--256.

\bibitem{levy_SeaThruNeRFNeuralRadiance_2023}
D.~Levy, A.~Peleg, N.~Pearl, D.~Rosenbaum, D.~Akkaynak, S.~Korman, and T.~Treibitz, ``{{SeaThru-NeRF}}: {{Neural Radiance Fields}} in {{Scattering Media}},'' in \emph{Proceedings of the {{IEEE}}/{{CVF Conference}} on {{Computer Vision}} and {{Pattern Recognition}}}, 2023, pp. 56--65.

\bibitem{sethuraman_WaterNeRFNeuralRadiance_2023}
A.~V. Sethuraman, M.~S. Ramanagopal, and K.~A. Skinner, ``{{WaterNeRF}}: {{Neural Radiance Fields}} for {{Underwater Scenes}},'' in \emph{{{OCEANS}} 2023 - {{MTS}}/{{IEEE U}}.{{S}}. {{Gulf Coast}}}, Sept. 2023, pp. 1--7.

\bibitem{qadri2022neural}
M.~Qadri, M.~Kaess, and I.~Gkioulekas, ``Neural implicit surface reconstruction using imaging sonar,'' in \emph{2023 IEEE International Conference on Robotics and Automation (ICRA)}.\hskip 1em plus 0.5em minus 0.4em\relax IEEE, 2023, pp. 1040--1047.

\bibitem{qadri2024aoneus}
\BIBentryALTinterwordspacing
M.~Qadri, K.~Zhang, A.~Hinduja, M.~Kaess, A.~Pediredla, and C.~A. Metzler, ``Aoneus: A neural rendering framework for acoustic-optical sensor fusion,'' in \emph{ACM SIGGRAPH 2024 Conference Papers}, ser. SIGGRAPH '24.\hskip 1em plus 0.5em minus 0.4em\relax New York, NY, USA: Association for Computing Machinery, 2024. [Online]. Available: \url{https://doi.org/10.1145/3641519.3657446}
\BIBentrySTDinterwordspacing

\bibitem{neusis-ngp}
Y.~Xie, G.~Troni, N.~Bore, and J.~Folkesson, ``Bathymetric surveying with imaging sonar using neural volume rendering,'' \emph{IEEE Robotics and Automation Letters}, vol.~9, no.~9, pp. 8146--8153, 2024.

\bibitem{dsc}
Y.~Feng, W.~Lu, H.~Gao, B.~Nie, K.~Lin, and L.~Hu, ``Differentiable space carving for 3{D} reconstruction using imaging sonar,'' \emph{IEEE Robotics and Automation Letters}, vol.~9, no.~11, pp. 10\,065--10\,072, 2024.

\bibitem{yiping_neural1}
\BIBentryALTinterwordspacing
Y.~Xie, N.~Bore, and J.~Folkesson, ``Sidescan only neural bathymetry from large-scale survey,'' \emph{Sensors}, vol.~22, no.~14, 2022. [Online]. Available: \url{https://www.mdpi.com/1424-8220/22/14/5092}
\BIBentrySTDinterwordspacing

\bibitem{kerbl3Dgaussians}
\BIBentryALTinterwordspacing
B.~Kerbl, G.~Kopanas, T.~Leimk{\"u}hler, and G.~Drettakis, ``3{D} gaussian splatting for real-time radiance field rendering,'' \emph{ACM Transactions on Graphics}, vol.~42, no.~4, July 2023. [Online]. Available: \url{https://repo-sam.inria.fr/fungraph/3d-gaussian-splatting/}
\BIBentrySTDinterwordspacing

\bibitem{yang2024seasplat}
D.~Yang, J.~J. Leonard, and Y.~Girdhar, ``Seasplat: Representing underwater scenes with 3d gaussian splatting and a physically grounded image formation model,'' in \emph{2025 IEEE International Conference on Robotics and Automation (ICRA)}, 2025.

\bibitem{li2024watersplatting}
H.~Li, W.~Song, T.~Xu, A.~Elsig, and J.~Kulhanek, ``Watersplatting: Fast underwater 3{D} scene reconstruction using gaussian splatting,'' in \emph{2025 International Conference on 3D Vision (3DV)}, 2025.

\bibitem{mualem2024gaussiansplashing}
N.~Mualem, R.~Amoyal, O.~Freifeld, and D.~Akkaynak, ``Gaussian splashing: Direct volumetric rendering underwater,'' 2024, \textit{arXiv:2411.19588}.

\bibitem{qu2024zsplatzaxisgaussiansplatting}
Z.~Qu, O.~Vengurlekar, M.~Qadri, K.~Zhang, M.~Kaess, C.~Metzler, S.~Jayasuriya, and A.~Pediredla, ``Z-splat: Z-axis gaussian splatting for camera-sonar fusion,'' \emph{IEEE Transactions on Pattern Analysis and Machine Intelligence}, pp. 1--12, 2024.

\bibitem{ge2022neural}
Y.~Ge, H.~Behl, J.~Xu, S.~Gunasekar, N.~Joshi, Y.~Song, X.~Wang, L.~Itti, and V.~Vineet, ``Neural-sim: Learning to generate training data with nerf,'' in \emph{Computer Vision -- ECCV 2022}, 2022, pp. 477--493.

\bibitem{loner}
S.~Isaacson, P.-C. Kung, M.~Ramanagopal, R.~Vasudevan, and K.~A. Skinner, ``Loner: Lidar only neural representations for real-time slam,'' \emph{IEEE Robotics and Automation Letters}, vol.~8, no.~12, pp. 8042--8049, 2023.

\bibitem{rosinol_NeRFSLAMRealTimeDense_2022}
A.~Rosinol, J.~J. Leonard, and L.~Carlone, ``Nerf-slam: Real-time dense monocular slam with neural radiance fields,'' in \emph{2023 IEEE/RSJ International Conference on Intelligent Robots and Systems (IROS)}, 2023, pp. 3437--3444.

\bibitem{ndf}
A.~Simeonov, Y.~Du, A.~Tagliasacchi, J.~B. Tenenbaum, A.~Rodriguez, P.~Agrawal, and V.~Sitzmann, ``Neural descriptor fields: Se(3)-equivariant object representations for manipulation,'' in \emph{2022 International Conference on Robotics and Automation (ICRA)}, 2022, pp. 6394--6400.

\bibitem{Matsuki:Murai:etal:CVPR2024}
H.~Matsuki, R.~Murai, P.~H. Kelly, and A.~J. Davison, ``Gaussian splatting slam,'' in \emph{Proceedings of the IEEE/CVF Conference on Computer Vision and Pattern Recognition (CVPR)}, June 2024, pp. 18\,039--18\,048.

\bibitem{luiten2023dynamic}
\BIBentryALTinterwordspacing
J.~Luiten, G.~Kopanas, B.~Leibe, and D.~Ramanan, ``{ Dynamic 3D Gaussians: Tracking by Persistent Dynamic View Synthesis },'' in \emph{2024 International Conference on 3D Vision (3DV)}.\hskip 1em plus 0.5em minus 0.4em\relax Los Alamitos, CA, USA: IEEE Computer Society, Mar. 2024, pp. 800--809. [Online]. Available: \url{https://doi.ieeecomputersociety.org/10.1109/3DV62453.2024.00044}
\BIBentrySTDinterwordspacing

\bibitem{radarfields}
\BIBentryALTinterwordspacing
D.~Borts, E.~Liang, T.~Broedermann, A.~Ramazzina, S.~Walz, E.~Palladin, J.~Sun, D.~Brueggemann, C.~Sakaridis, L.~Van~Gool, M.~Bijelic, and F.~Heide, ``Radar fields: Frequency-space neural scene representations for fmcw radar,'' in \emph{ACM SIGGRAPH 2024 Conference Papers}, ser. SIGGRAPH '24.\hskip 1em plus 0.5em minus 0.4em\relax New York, NY, USA: Association for Computing Machinery, 2024. [Online]. Available: \url{https://doi.org/10.1145/3641519.3657510}
\BIBentrySTDinterwordspacing

\bibitem{cloner}
A.~Carlson, M.~S. Ramanagopal, N.~Tseng, M.~Johnson-Roberson, R.~Vasudevan, and K.~A. Skinner, ``Cloner: Camera-lidar fusion for occupancy grid-aided neural representations,'' \emph{IEEE Robotics and Automation Letters}, vol.~8, no.~5, pp. 2812--2819, 2023.

\bibitem{huang2024dartimplicitdopplertomography}
T.~Huang, J.~Miller, A.~Prabhakara, T.~Jin, T.~Laroia, Z.~Kolter, and A.~Rowe, ``Dart: Implicit doppler tomography for radar novel view synthesis,'' in \emph{Proceedings of the IEEE/CVF Conference on Computer Vision and Pattern Recognition (CVPR)}, June 2024, pp. 24\,118--24\,129.

\bibitem{lin2024thermalnerf}
\BIBentryALTinterwordspacing
Y.~Y. Lin, X.-Y. Pan, S.~Fridovich-Keil, and G.~Wetzstein, ``{ ThermalNeRF: Thermal Radiance Fields },'' in \emph{2024 IEEE International Conference on Computational Photography (ICCP)}.\hskip 1em plus 0.5em minus 0.4em\relax Los Alamitos, CA, USA: IEEE Computer Society, July 2024, pp. 1--12. [Online]. Available: \url{https://doi.ieeecomputersociety.org/10.1109/ICCP61108.2024.10644336}
\BIBentrySTDinterwordspacing

\bibitem{wide-apeture-reconstruction}
\BIBentryALTinterwordspacing
T.~Guerneve, K.~Subr, and Y.~Petillot, ``Three-dimensional reconstruction of underwater objects using wide-aperture imaging sonar,'' \emph{Journal of Field Robotics}, vol.~35, no.~6, pp. 890--905, 2018. [Online]. Available: \url{https://onlinelibrary.wiley.com/doi/abs/10.1002/rob.21783}
\BIBentrySTDinterwordspacing

\bibitem{valdenegro2016submerged}
M.~Valdenegro-Toro, ``Submerged marine debris detection with autonomous underwater vehicles,'' in \emph{2016 International Conference on Robotics and Automation for Humanitarian Applications (RAHA)}.\hskip 1em plus 0.5em minus 0.4em\relax IEEE, 2016, pp. 1--7.

\bibitem{stereo_FS}
S.~Negahdaripour, ``On 3-{D} reconstruction from stereo {FS} sonar imaging,'' in \emph{OCEANS 2010 MTS/IEEE SEATTLE}, 2010, pp. 1--6.

\bibitem{Potokar22iros}
E.~Potokar, K.~Lay, K.~Norman, D.~Benham, T.~B. Neilsen, M.~Kaess, and J.~G. Mangelson, ``Holoocean: Realistic sonar simulation,'' in \emph{2022 IEEE/RSJ International Conference on Intelligent Robots and Systems (IROS)}, 2022, pp. 8450--8456.

\bibitem{wang20242dforwardlookingsonar}
Y.~Wang, C.~Wu, Y.~Ji, H.~Tsuchiya, H.~Asama, and A.~Yamashita, ``2{D} forward looking sonar simulation with ground echo modeling,'' in \emph{2023 20th International Conference on Ubiquitous Robots (UR)}, 2023, pp. 724--729.

\bibitem{afm_kaess}
T.~A. Huang and M.~Kaess, ``Towards acoustic structure from motion for imaging sonar,'' in \emph{2015 IEEE/RSJ International Conference on Intelligent Robots and Systems (IROS)}, 2015, pp. 758--765.

\bibitem{albedo_kaess}
E.~Westman, I.~Gkioulekas, and M.~Kaess, ``A volumetric albedo framework for 3{D} imaging sonar reconstruction,'' in \emph{2020 IEEE International Conference on Robotics and Automation (ICRA)}, 2020, pp. 9645--9651.

\bibitem{lihigs}
\BIBentryALTinterwordspacing
P.-C. Kung, X.~Zhang, K.~A. Skinner, and N.~Jaipuria, ``Lihi-gs: Lidar-supervised gaussian splatting for highway driving scene reconstruction,'' 2024, \textit{arXiv:2412.15447}. [Online]. Available: \url{https://arxiv.org/abs/2412.15447}
\BIBentrySTDinterwordspacing

\bibitem{lee2025microsplattingmaximizingisotropicconstraints}
\BIBentryALTinterwordspacing
J.~W. Lee, H.~Lim, S.~Yang, and J.~Choi, ``Micro-splatting: Maximizing isotropic constraints for refined optimization in 3d gaussian splatting,'' 2025. [Online]. Available: \url{https://arxiv.org/abs/2504.05740}
\BIBentrySTDinterwordspacing

\bibitem{ye2024gsplatopensourcelibrarygaussian}
\BIBentryALTinterwordspacing
V.~Ye, R.~Li, J.~Kerr, M.~Turkulainen, B.~Yi, Z.~Pan, O.~Seiskari, J.~Ye, J.~Hu, M.~Tancik, and A.~Kanazawa, ``gsplat: An open-source library for {Gaussian} splatting,'' \emph{arXiv preprint arXiv:2409.06765}, 2024. [Online]. Available: \url{https://arxiv.org/abs/2409.06765}
\BIBentrySTDinterwordspacing

\bibitem{subsea_blueprint_nodate}
\BIBentryALTinterwordspacing
``\BIBforeignlanguage{en}{Blueprint {Subsea} {\textbar} {Oculus} {M}-{Series}}.'' [Online]. Available: \url{https://www.blueprintsubsea.com/}
\BIBentrySTDinterwordspacing

\bibitem{noauthor_dvl_nodate}
\BIBentryALTinterwordspacing
``\BIBforeignlanguage{en-US}{{DVL} {A50}}.'' [Online]. Available: \url{https://waterlinked.com/shop/dvl-a50-114}
\BIBentrySTDinterwordspacing

\bibitem{leroy2024groundingimagematching3d}
\BIBentryALTinterwordspacing
V.~Leroy, Y.~Cabon, and J.~Revaud, ``Grounding image matching in 3d with mast3r,'' 2024. [Online]. Available: \url{https://arxiv.org/abs/2406.09756}
\BIBentrySTDinterwordspacing

\bibitem{shen2008map}
H.~Shen and L.~Zhang, ``A map-based algorithm for destriping and inpainting of remotely sensed images,'' \emph{IEEE Transactions on Geoscience and Remote Sensing}, vol.~47, no.~5, pp. 1492--1502, 2008.

\bibitem{rakwatin2007stripe}
P.~Rakwatin, W.~Takeuchi, and Y.~Yasuoka, ``Stripe noise reduction in modis data by combining histogram matching with facet filter,'' \emph{IEEE Transactions on Geoscience and Remote Sensing}, vol.~45, no.~6, pp. 1844--1856, 2007.

\bibitem{nichol2004noise}
J.~E. Nichol and V.~Vohora, ``Noise over water surfaces in landsat tm images,'' \emph{International Journal of Remote Sensing}, vol.~25, no.~11, pp. 2087--2093, 2004.

\bibitem{Sethuraman_2023_BMVC}
\BIBentryALTinterwordspacing
A.~V. Sethuraman and K.~A. Skinner, ``Stars: Zero-shot sim-to-real transfer for segmentation of shipwrecks in sonar imagery,'' in \emph{34th British Machine Vision Conference 2023, {BMVC} 2023, Aberdeen, UK, November 20-24, 2023}.\hskip 1em plus 0.5em minus 0.4em\relax BMVA, 2023. [Online]. Available: \url{https://papers.bmvc2023.org/0606.pdf}
\BIBentrySTDinterwordspacing

\bibitem{gmvatr}
A.~V. Sethuraman, P.~Baldoni, K.~A. Skinner, and J.~McMahon, ``Learning which side to scan: Multi-view informed active perception with side scan sonar for autonomous underwater vehicles,'' in \emph{2024 IEEE International Conference on Robotics and Automation (ICRA)}, 2024, pp. 8348--8354.

\end{thebibliography}

\end{document}